\documentclass[letterpaper, 10 pt, conference]{ieeeconf}  

\IEEEoverridecommandlockouts                              

\usepackage{times}
\usepackage{soul}
\usepackage{url}
\usepackage[hidelinks]{hyperref}
\usepackage[utf8]{inputenc}
\usepackage[small]{caption}
\usepackage{graphicx}
\usepackage{amsmath}
\usepackage{amsthm}
\usepackage{booktabs}
\usepackage{algorithm}
\usepackage{algorithmic}
\usepackage[switch]{lineno}
\usepackage{graphics} 
\usepackage{graphicx}
\usepackage{amssymb}  
\usepackage{cleveref}
\usepackage{caption}
\usepackage{siunitx}
\usepackage[font=footnotesize]{caption}
\usepackage{float}
\usepackage{subfig}
\usepackage[dvipsnames]{xcolor}
\usepackage{color, colortbl}

\title{\LARGE \bf
Learning global control of underactuated systems with Model-Based Reinforcement Learning
}

\author{Niccolò Turcato$^1$, Marco Calì$^1$, Alberto Dalla Libera$^1$,  Giulio Giacomuzzo$^1$, Ruggero Carli$^1$ and Diego Romeres$^2$
\thanks{}
\thanks{$^{1}$Department of Information Engineering, University of Padova, Italy.}%
\thanks{$^{2}$Mitsubishi Electric Research Laboratories, Cambridge, MA, USA}%
\thanks{Correspondence to {\tt\small turcatonic@dei.unipd.it}}%
}

\begin{document}

\definecolor{LightRed}{rgb}{1,0.7,0.7}
\newcommand{\R}{\mathbb{R}}
\newcommand{\q}{\boldsymbol{q}}
\newcommand{\dq}{\boldsymbol{\dot{q}}}
\newcommand{\ddq}{\boldsymbol{\ddot{q}}}
\newcommand{\taubf}{\boldsymbol{\tau}}
\newcommand{\cor}{\boldsymbol{c}}
\newcommand{\X}{\boldsymbol{X}}
\newcommand{\x}{\boldsymbol{x}}
\newcommand{\ub}{\boldsymbol{u}}
\newcommand{\ab}{\boldsymbol{a}}
\newcommand{\thetab}{\boldsymbol{\theta}}
\newcommand{\Y}{\boldsymbol{Y}}
\newcommand{\y}{\boldsymbol{y}}
\newcommand{\w}{\boldsymbol{w}}
\newcommand{\p}{\boldsymbol{p}}
\newcommand{\Pb}{\boldsymbol{P}}
\newcommand{\Deltab}{\boldsymbol{\Delta}}
\newcommand{\deltab}{\boldsymbol{\delta}}
\newcommand{\Lag}{\mathcal{L}}
\newcommand{\E}{\mathbb{E}}
\newcommand{\norm}[1]{\left\lVert#1\right\rVert}
\newcommand{\D}{\mathcal{D}}
\newcommand{\f}{\mathbf{f}}
\newcommand{\e}{\boldsymbol{e}}
\newcommand{\mub}{\boldsymbol{\mu}}
\newcommand{\Fb}{\boldsymbol{F}}
\newcommand{\TODO}{\textcolor{red}{\textbf{TODO}}}

\maketitle
\thispagestyle{empty}
\pagestyle{empty}

\begin{abstract}
This short paper describes our proposed solution for the third edition of the "AI Olympics with RealAIGym" competition, held at ICRA 2025.
We employed Monte-Carlo Probabilistic Inference for Learning Control (MC-PILCO), an MBRL algorithm recognized for its exceptional data efficiency across various low-dimensional robotic tasks, including cart-pole, ball \& plate, and Furuta pendulum systems. MC-PILCO optimizes a system dynamics model using interaction data, enabling policy refinement through simulation rather than direct system data optimization. This approach has proven highly effective in physical systems, offering greater data efficiency than Model-Free (MF) alternatives. Notably, MC-PILCO has previously won the first two editions of this competition, demonstrating its robustness in both simulated and real-world environments. 
Besides briefly reviewing the algorithm, we discuss the most critical aspects of the MC-PILCO implementation in the tasks at hand: learning a global policy for the pendubot and acrobot systems.

\end{abstract}


\section{Introduction}
This report outlines our team's implementation of a Reinforcement Learning (RL) approach to address the third "AI Olympics with RealAIGym" competition at ICRA 2025\footnote{\url{https://ai-olympics.dfki-bremen.de/}}, based on the RealAIGym project \cite{wiebe2022realaigym}. Specifically, we employed Monte-Carlo Probabilistic Inference for Learning Control (MC-PILCO) \cite{amadio2021mc_pilco}, a Model-Based (MB) RL algorithm known for its exceptional data efficiency in various low-dimensional benchmarks, including cart-pole, ball \& plate, and Furuta pendulum systems, both in simulation and real-world environments. Notably, MC-PILCO also secured victory in the first two editions of this competition \cite{ijcai2024p1043, wiebe2025reinforcement, turcato2024learningcontrol}.
MC-PILCO leverages interaction data to optimize a system dynamics model. Instead of directly optimizing the policy using system data, it refines the policy by simulating the system, enhancing data efficiency.
Considering physical systems, this approach can be highly performing and more data-efficient than Model-Free (MF) solutions.
Examples of MC-PILCO applications and derivations have been reported in \cite{amadio2023mcpilco_raw_meas,mcpilco_tossing,turcato2025dataefficientroboticobject,LIANG2024394}. 

This paper is organized as follows: \Cref{sec:goal} introduces the goal and the settings of the competition. \Cref{sec:methods} presents the MC-PILCO algorithm for global policy training. \Cref{sec:experiments} reports the experiments that have been performed, finally \Cref{sec:conclusions} concludes the paper.

Compared to the solutions proposed in the first two editions, the solution proposed for this competition integrates a new type of incremental training (\Cref{subsec:global_training}), which aims at developing a global controller for the system.

\section{Competition Overview}
\label{sec:goal}
This challenge focuses on a two-degrees-of-freedom (2-DoF) underactuated pendulum system, as described in \cite{10375556}, which can be set in two configurations. In the first configuration—known as the Pendubot—the joint connected to the base is actuated, while the second joint is passive. In the second configuration—referred to as the Acrobot—the first joint is passive, and the second is actuated. For both configurations, the objective is to design a controller capable of performing swing-up and stabilizing the pendulum at its unstable equilibrium point. Due to the underactuated nature of both systems, this task presents significant control challenges.

The systems are simulated using a fourth-order Runge-Kutta integrator at a rate of 500 Hz over a time horizon of $T=\SI{60}{\s}$. The competition is structured in two stages. The first stage—the simulation stage—evaluates the controllers in a simulated environment. In the second stage—the hardware stage—teams test their controllers on the physical system, with the option to retrain learning-based approaches.

Since the ultimate goal is to develop a global policy, controllers are tested by randomly initializing the system from various points in the state space at random times. The competition winners are determined based on both the performance and reliability of their submitted controllers.


\section{Learning a global policy with MC-PILCO}
\label{sec:methods}
This section first reviews MC-PILCO, secondly it discusses its application to learn a global policy for the underactuated double pendulum. 

\subsection{MC-PILCO}
\label{subsec:mc_pilco_review}
MC-PILCO is a Model-Based policy gradient algorithm. Gaussian Processes (GPs) are used to estimate system dynamics and long-term state distributions are approximated with a particle-based method.

Consider a system with evolution described by the discrete-time unknown transition function $f: \R^{d_x} \times \R^{d_u} \rightarrow \R^{d_x}$:
\begin{equation}
    \x_{t+1} = f(\x_{t}, \ub_{t}) + \w_{t},
    \label{eq:system_equation}
\end{equation}
where $\x_{t} \in \R^{d_x}$ and $\ub_{t} \in \R^{d_u}$ are respectively the state and input of the system at step $t$, while $\w_{t}$ is an independent white noise describing uncertainty influencing the system evolution. As usual in RL, a cost function $c(\x_{t})$ encodes the task to be accomplished. A policy $\pi_{\thetab}: \x \rightarrow \ub$ that depends on the parameters $\thetab$ selects the inputs applied to the system. The objective is to find policy parameters $\thetab^*$ that minimize the cumulative expected cost, defined as follows,
\begin{equation}
    J(\thetab) = \sum_{t=0}^T \E [c(\x_{t})],
    \label{eq:cumulative_cost}
\end{equation}
where the initial state $x_0$ is sampled according to a given probability $p(\x_0)$.

MC-PILCO consists of a series of attempts, known as trials, to solve the desired task.
Each trial consists of three main phases: (i) model learning, (ii) policy update, and (iii) policy execution. In the first trial, the GP model is derived from data collected with an exploration policy, for instance, a random exploration policy. 

In the model learning step, previous experience is used to build or update a model of the system dynamics. The policy update step formulates an optimization problem whose objective is to minimize the cost in \cref{eq:cumulative_cost} w.r.t. the parameters of the policy $\thetab$. 
Finally, in the last step, the current optimized policy is applied to the system and the collected samples are stored to update the model in the next trials.

In the rest of this section, we give a brief overview of the main components of the algorithm and highlight their most relevant features.

\subsubsection{Model Learning}
\label{subsubsec:model_learning}
MC-PILCO relies on GP Regression (GPR) to learn the system dynamics \cite{rasmussen2003gps_for_ml}. For the use of GPs in system identification and control we refer the interested reader to
\cite{kernel_methods_and_gp_control_systems_magazine}.
In our previous work, \cite{amadio2021mc_pilco}, we presented a framework specifically designed for mechanical systems, named speed-integration model. 
Given a mechanical system with $d$ degrees of freedom, the state is defined as $\x_t = [\q_t^T, \dq_t^T]^T$ where $\q_t \in \mathbb{R}^d$ and $\dq_t \in \mathbb{R}^d$ are, respectively, the generalized positions and velocities of the system at time $t$. 
Let $T_s$ be the sampling time and assume that accelerations between successive time steps are constant. 
The following equations describe the one-step-ahead evolution of the $i$-th degree of freedom, 

\begin{subequations} \label{eq:speed-int}
\begin{align}
    &\dot{q}^{(i)}_{t+1} = \dot{q}^{(i)}_{t} + \Delta ^{(i)}_t \label{eq:speed-int-vel}\\
    &q_{t + 1}^{(i)} = q_t^{(i)} + T_s \dot{q}^{(i)}_t + \frac{T_s}{2} \Delta ^{(i)}_t \label{eq:speed-pos}
\end{align}
\end{subequations}
where $\Delta ^{(i)}_t$ is the change in velocity. MC-PILCO estimates the unknown function $\Delta^{(i)}_t$ from collected data by GPR. Each $\Delta_t^{(i)}$ is modeled as an independent GP, denoted $f^i$, with input vector $\tilde{\x}_t = [\x^T_t, \ub^T_t]^T$, hereafter referred as GP input. 

The posterior distributions of each $\Delta_t^{(i)}$ given $\mathcal{D}^{i}$ are Gaussian distributed, with mean and variance expressed as follows:
\begin{equation}
    \begin{split}
        &\E[\hat{\Delta}_t^{(i)}] = m^{(i)}_\Delta(\tilde{\x}_t) + K_{\tilde{\x}_t \tilde{\X}} \Gamma_i^{-1} (\y^{(i)} - m^{(i)}_\Delta(\tilde{\X})),    \\
        &var[\hat{\Delta}_t^{(i)}] = k_i(\tilde{\x}_t, \tilde{\x}_t) - K_{\tilde{\x}_t \tilde{\X}} \Gamma_i^{-1} K_{\tilde{\X} \tilde{\x}_t}, \\
        &\Gamma_i = K_{\tilde{\X} \tilde{\X}} + \sigma_i^2 I,
    \end{split}
    \label{eq:gp_regression_formulas}
\end{equation}
refer to \cite{rasmussen2003gps_for_ml} for the derivation of \Cref{eq:gp_regression_formulas}.
Then, also the posterior distribution of the one-step ahead transition model in \eqref{eq:speed-int} is Gaussian, namely, 
\begin{equation}
    p(\x_{t+1} | \x_t, \ub_t, \D) \sim \mathcal{N}(\mub_{t+1}, \Sigma_{t+1}) 
    \label{eq:one-step-posteior}
\end{equation}
with mean $\mub_{t+1}$ and covariance $\Sigma_{t+1}$ derived combining \eqref{eq:speed-int} and \eqref{eq:gp_regression_formulas}.

\subsubsection{Policy Update}
\label{subsubsec:policy_update}
In the policy update phase, the policy is trained to minimize the expected cumulative cost in \cref{eq:cumulative_cost} with the expectation computed w.r.t. the one-step ahead probabilistic model in \cref{eq:one-step-posteior}. This requires the computation of long-term distributions starting from the initial distribution $p(\x_0)$ and \cref{eq:one-step-posteior}, which is not possible in closed form. MC-PILCO resorts to Monte Carlo sampling \cite{caflisch1998monte_carlo_sampling_ref} to approximate the expectation in \cref{eq:cumulative_cost}. The Monte Carlo procedure starts by sampling from $p(\x_0)$ a batch of $N$ particles and simulates their evolution based on the one-step-ahead evolution in \cref{eq:one-step-posteior} and the current policy. Then, the expectations in \cref{eq:cumulative_cost} are approximated by the mean of the simulated particles costs, namely,

\begin{equation}
    \begin{split}
        \hat{J}(\boldsymbol{\thetab}) = \sum_{t=0}^{T} \left( \frac{1}{N} \sum_{n=1}^N c \left( \x_t^{(n)} \right) \right) 
    \end{split}
    \label{eq:cost_estimate_monte_carlo}
\end{equation}
where $\x_t^{(n)}$ is the state of the $n$-th particle at time $t$.

The optimization problem is interpreted as a stochastic gradient descend problem (SGD) \cite{bottou2010large_scale_learning_sgd}, applying the reparameterization trick to differentiate stochastic operations \cite{kingma2013reparametrization_trick}.

The authors of \cite{amadio2021mc_pilco} proposed the use of dropout \cite{srivastava2014dropout} of the policy parameters $\thetab$ to improve exploration and increase the ability to escape from local minima during policy optimization of MC-PILCO.

\subsection{Global Policy training}
\label{subsec:global_training}
The task in object presents several practical issues when applying the algorithm. The first one is that the control frequency requested by the challenge is quite high for a MBRL approach. Indeed, high control frequencies require a high number of model evaluations which increases the computational cost of the algorithm. 
The double pendulum system from the RealAIGym project can be controlled at relatively low frequencies like similar systems \cite{amadio2021mc_pilco, amadio2023learning}. 
Indeed, in the real hardware stage of the first two editions of the competition, the MC-PILCO controller was trained to work at \SI{33}{\hertz} \cite{ijcai2024p1043, wiebe2025reinforcement}.
However, the absence of friction in the simulated system makes the system particularly sensitive to the system input. Hence, we selected a control frequency of \SI{50}{\hertz} for this stage.

The second challenge lies in the task requirements. Indeed, the task requires the policy to drive the system to the unstable equilibrium starting from an initial state $x_0=[p_0^T, \dot{p}_0^T]^T$, where $p_0 \in [-\pi, \pi] \times [-\pi,\pi]$ and $||\dot{p}_0||$ is very small.
Thus the initial state distribution of the system can be defined as
\begin{equation}
    p(x_0) \sim U\left(-x_M, x_M \right), x_M =\begin{bmatrix}
        \pi \\
        \pi \\
        \varepsilon \\
        \varepsilon
    \end{bmatrix} 
    \label{eq:initial_state}
\end{equation}
where $\varepsilon > 0$ is a very small constant.

Since the nominal model of the system is available to develop the controller, we use the forward dynamics function of the plant as the prior mean function of the change in velocity for each joint. The available model is
\begin{equation}
    B \ub_t = M(\q_t) \ddq_t + n(\q_t,\dq_t),
\end{equation}
where $M(\q_t)$ is the mass matrix, $n(\q_t,\dq_t)$ contains the Coriolis, gravitational and damping terms, and $B$ is the actuation matrix, which is $B=\text{diag}([1, 0])$ for the Pendubot and $B=\text{diag}([0, 1])$ for the Acrobot.
We define then
\begin{equation}
    m_\Delta(\tilde{\x}_t) = \begin{bmatrix} m_\Delta^{(1)} \\ m_\Delta^{(2)}\end{bmatrix} := T_s \cdot M^{-1}(\q_t) (B \ub_t - n(\q_t,\dq_t))
    \label{eq:mean_fun_forward_dyn}
\end{equation}
as the mean function in \cref{eq:gp_regression_formulas}. The control input $\ub_t \in \R$ is a scalar representing the torque given in input to the controlled joint.
It is important to point out that \cref{eq:mean_fun_forward_dyn} is nearly perfect to approximate the system when $T_s$ is sufficiently small, but it becomes unreliable as $T_s$ grows. In particular, with $T_s=\SI{0.02}{\s}$ the predictions of \cref{eq:mean_fun_forward_dyn} are not good enough to describe the behavior at the unstable equilibrium. The inaccuracies of the prior mean are compensated by the GP models. To cope with the large computational burden due to the high number of collected samples, we implemented the GP approximation Subset of Data, see \cite{JMLR:v6:quinonero-candela05a} for a detailed description. 

An important aspect of policy optimization is the particle initialization, in this case using the initial distribution \cref{eq:initial_state} in \cref{eq:cost_estimate_monte_carlo} at the first trial results in a very unreliable optimization problem, which typically does not converge to acceptable solutions.
For this reason, we employ an incremental initialization strategy to learn global control for the system. Namely we use a surrogate initial distribution for both policy execution and particles initialization:
\begin{align}
    p'_k(x_0) &\sim U \left( - x_M \cdot \gamma_k, x_M \cdot \gamma_k \right),\\ 
    \gamma_k &= \text{clip}(\frac{k-k_m}{K}, 0, 1),
    \label{eq:surrogate_init}
\end{align}
where $k \in \mathbb{N}$ is the trial index, and $k_m, K \in \mathbb{N}$ regulate the increment in the uniform distribution's bounds. This strategy falls within Curriculum Learning \cite{9392296}, as the policy is trained on a task of increasing difficulty.
\begin{figure}
    \centering
    \includegraphics[width=\linewidth]{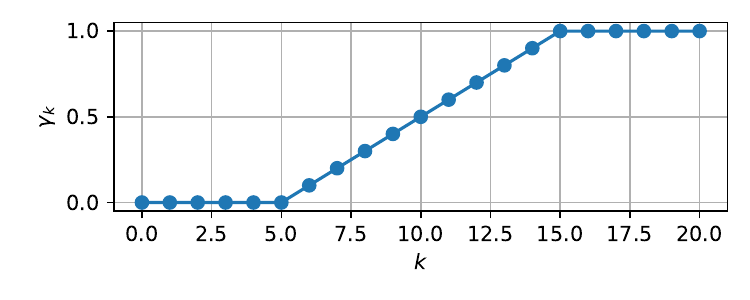}
    \caption{$\gamma_k$ scheduling following \cref{eq:surrogate_init}, with $k_m=5, K=10$.}
    \label{fig:gamma_scheduling}
\end{figure}

The cost function must evaluate the policy performance w.r.t. the task requirements, in this case, we want the system to reach the position $\q_G=[\pi, 0]^T$ and stay there indefinitely. A cost generally used in this kind of system is the saturated distance from the target state:
\begin{align}
    \begin{split}
        c_{st}(\x_t) = 1 - e^{- \norm{|\q_t| - \q_G}^2_{\Sigma_c}}   \hspace{0.5cm}
        \Sigma_c = \text{diag}\left(\frac{1}{\ell_c}, \frac{1}{\ell_c}\right),
    \end{split}
    \label{eq:saturated_dist_target}
\end{align}
with $\ell_c=3$. Notice that this cost does not depend on the velocity of the system, just on the distance from the goal state, but it does encourage the policy to reach the goal state with zero velocity.

The policy function that is used to learn a control strategy is the general purpose policy from \cite{amadio2021mc_pilco}:
\begin{equation}
    \begin{split}
         &\pi_{\thetab}(\x_t) = u_M \tanh{ \left(  \sum_{i=1}^{N_b} \frac{w_i}{u_{M}} e^{-\| \ab_i - \phi(\x_t) \|^2_{\Sigma_\pi}}  \right)} \\
         &\phi(\x_t) = [\dq_t^T, \cos{(\q_t^T)}, \sin{(\q_t^T)}]^T\\
    \end{split}
    \label{eq:tossing_policy}
\end{equation}

with hyperparameters $\thetab = \{{\bf w}, A, \Sigma_\pi\}$, where ${\bf w} = [w_1, \dots, w_{N_b}]^T$ and $A= \{\ab_1, \dots, \ab_{N_b}\}$ are, respectively, weights and centers of the $N_b$ Gaussians basis functions, whose shapes are determined by $\Sigma_{\pi}$. For both robots, the dimensions of the elements of the policy are: $\Sigma_{\pi} \in \R^{6 \times 6}$, $a_i \in \R^6$, $w_i \in \R$ for $i = 1, \dots, N_b$, since the policy outputs a single scalar. In the experiments, the parameters are initialized as follows. The basis weights are sampled uniformly in $[-u_M, u_M]$, the centers are sampled uniformly in the image of $\phi$ with $\dq_t \in [-2 \pi, 2 \pi]$ rad/s. The matrix $\Sigma_{\pi}$ is initialized to the identity. 


\begin{figure}
    \centering
    \includegraphics[width=\linewidth]{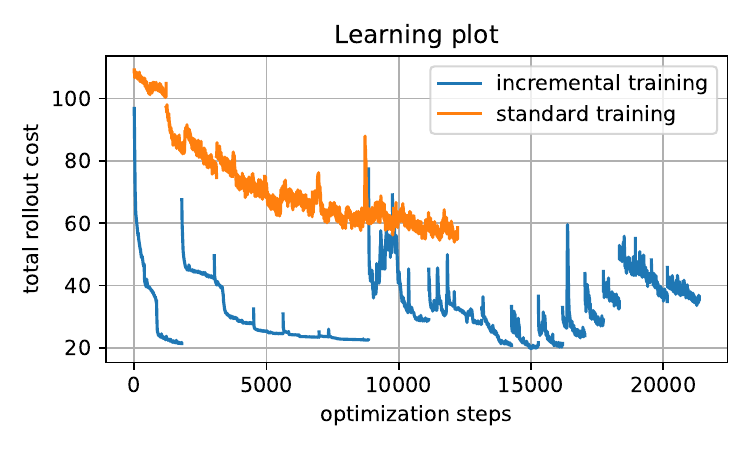}
    \caption{Total rollout costs in the policy optimization steps of the two MC-PILCO trainings, the first using the incremental initial distribution, the second using the nominal initial distribution in all trials.}
    \label{fig:learning_plot}
\end{figure}

\section{Experiments}
\label{sec:experiments}
In this section, we briefly discuss how the algorithm was applied to both systems and show the main results. We also report the optimization parameters used for both systems, all the parameters not specified are set to the values reported in \cite{amadio2021mc_pilco}.
All the code was implemented in Python with the PyTorch \cite{paszke2017automatic_diff_pytorch} library.

For both robots, we use the model described in \Cref{subsubsec:model_learning}, with mean function from \cref{eq:mean_fun_forward_dyn} and squared-exponential kernel \cite{rasmussen2003gps_for_ml}.
The policy optimization horizon was set much lower than the horizon required for the competition, this allows to reduce the computational burden of the algorithm, moreover, it pushes the optimization to find policies that can execute a fast swing-up. We exploit dropout in the policy optimization as a regularization strategy, to yield better policies.

\emph{Note:} The performance score for the simulation stage is computed by simulating the controllers for $\SI{60}{\s}$, randomly resetting the joints to some position in $[-\pi, \pi]$. The score is proportional to the duration for which the system remains stabilized in the unstable equilibrium\footnote{\url{https://ai-olympics.dfki-bremen.de/}}. However, this is done with a PID controller activated for $\SI{0.2}{\s}$, which, when the joints are outside $[-\pi, \pi]$ causes the system to accelerate to high velocity, becoming uncontrollable, due to the lack of friction. This can be partially solved on the pendubot by switching to a damping control when the joint velocity is too high, while it remains unsolved for the acrobot.

\subsection{Pendubot}
\label{sec:pendubot}

\begin{figure}
    \centering	
    \includegraphics[width=\columnwidth]{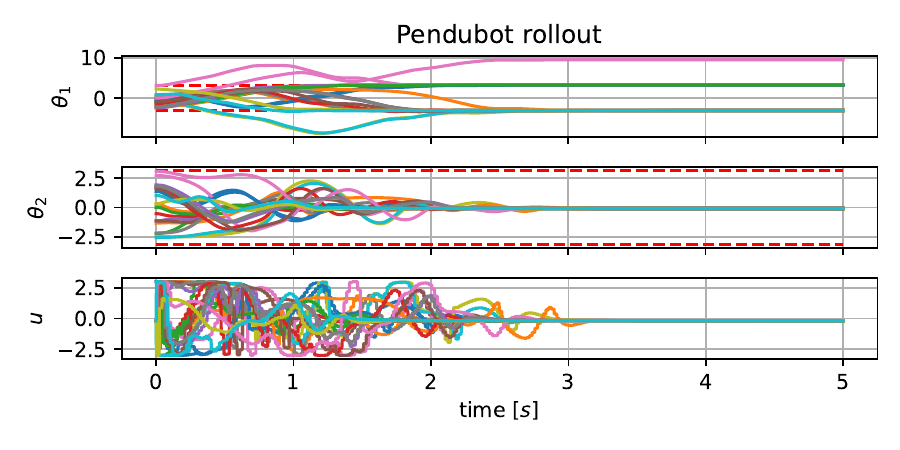}
    \caption{20 simulated trials of the Pendubot system (\SI{500}{\hertz}), under MC-PILCO's control policy (\SI{50}{\hertz}). The initial position for each joint is uniformly sampled from the interval $[-\pi,\pi]$.}
    \label{fig:pendubot_rollout}
\end{figure}

The policy for the Pendubot swing-up was optimized for a horizon of $T=\SI{3.0}{\s}$, with $u_M=\SI{3.0}{\N\cdot\m}$. The control switches to a damping controller $\ub = - D \dq_1$ when $\text{max}(\dq_1, \dq_2) \geq \SI{20}{rad / \s}$ to limit uncontrollable maneuvers caused by the resetting PID. The Linear Quadratic Regulator (LQR) controller for stabilization is not used for this system, to avoid additional complexity in the control strategy.
The policy is optimized over a total of $20$ episodes, using $\gamma_k$ scheduling reported in \cref{fig:gamma_scheduling} ($k_m=5, K=10$).

In \cref{fig:learning_plot} we compare the total rollout costs of the policy update steps obtained with the proposed incremental training, with the standard approach using the nominal initial distribution \cref{eq:initial_state} in all trials.
While the number of episodes of the two trainings is the same, the number of steps in the optimizations is quite different. Specifically, the incremental training allows the policy to first learn the swing-up task, starting from the stable equilibrium, and then gradually adapt the parameters to a wider initial distribution. 
In contrast, using the nominal uniform distribution since the first trial results in a much more complex optimization, because the initial parameters of the policy are random and the model's prediction is more uncertain (having less data). This results in noisy policy gradient steps, which trigger the exit condition from the optimization sooner than with the proposed strategy. As a result, the final total rollout cost of the policy trained with the incremental initial distribution is lower than with the standard training.

The Controller's strategy is depicted in \cref{fig:pendubot_rollout}. This controller has a performance score close to $0.5$, while the MC-PILCO controller obtained with the standard training achieves a score of $0.1$. The baseline controller Time Varing Linear Quadratic Regulator (TVLQR) \cite{ijcai2024p1043,10375556} has a score $\leq 0.1$. 
\Cref{tab:perf_table} reports the scores of our controller and the baseline.

\subsection{Acrobot}
\label{sec:acrobot}
The policy for the Acrobot swing-up was optimized for a horizon of $T=\SI{2.0}{\s}$, with $u_M=\SI{3.0}{\N\cdot\m}$. 
The policy is optimized over a total of $20$ episodes, using $\gamma_k$ scheduling reported in \cref{fig:gamma_scheduling} ($k_m=5, K=10$).
The optimization steps for the acrobot's policy result in a learning plot similar to \cref{fig:learning_plot}.

Given the ideal conditions considered in this simulation, the control switches to an LQR controller after the swing-up.
Under ideal circumstances, the LQR controller can stabilize the system at an unstable equilibrium by exerting zero final torque. The switching condition is obtained by checking if the system's state is within the LQR's region of attraction.

\begin{figure}
    \centering	
    \includegraphics[width=\columnwidth]{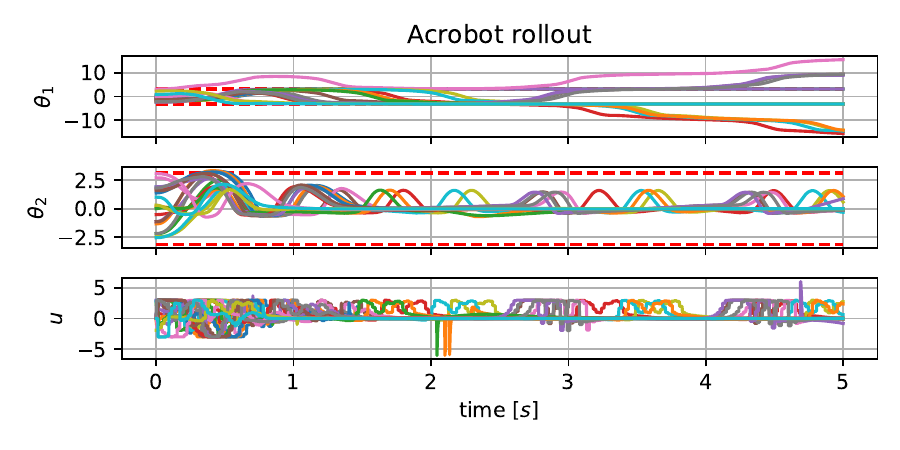}
    \caption{20 simulated trials of the Pendubot system (\SI{500}{\hertz}), under MC-PILCO's control policy (\SI{50}{\hertz}). The initial position for each joint is uniformly sampled from the interval $[-\pi,\pi]$.}
    \label{fig:acrobot_rollout}
\end{figure}

The Controller's strategy is depicted in \cref{fig:acrobot_rollout}. This controller has a performance score of around $0.3$, while with standard training it is slightly lower, around $0.2$.
The baseline TVLQR has a score lower than $0.1$. 
\Cref{tab:perf_table} reports the scores of our controller and the baseline. 

\begin{center}
\begin{table}[]
\begin{tabular}{|| c || c || c ||} 
 \hline
 Controller & Perf. score Pendubot & Perf. score Acrobot \\ [0.5ex] 
 \hline \hline
\rowcolor{LightRed} MC-PILCO (incremental) & 0.468 & 0.292 \\
 \hline
MC-PILCO (standard) & 0.1 & 0.21 \\
 \hline
TVLQR & 0.094 & 0.073 \\
\hline
\end{tabular}
\caption{Pendubot and Acrobot scores comparison.}
\label{tab:perf_table}
\end{table}
\end{center}

\section{Conclusions}
\label{sec:conclusions}
In both systems, our MBRL approach is able to solve the global task with good swing-up time, handling the uniform initial state distribution.

\section*{Acknowledgements}
Alberto Dalla Libera and Giulio Giacomuzzo were supported by PNRR research activities of the consortium iNEST (Interconnected North-Est Innovation Ecosystem) funded by the European Union Next GenerationEU (Piano Nazionale di Ripresa e Resilienza (PNRR) – Missione 4 Componente 2, Investimento 1.5 – D.D. 1058  23/06/2022, ECS\_00000043). This manuscript reflects only the Authors’ views and opinions, neither the European Union nor the European Commission can be considered responsible for them.

\bibliographystyle{ieeetr}
\bibliography{ref}

\end{document}